%% file: main.tex
\documentclass[11pt]{article}

\usepackage{acl}

\usepackage{times}
\usepackage{latexsym}
\usepackage[T1]{fontenc}
\usepackage[utf8]{inputenc}
\usepackage{microtype}
\usepackage{dirtytalk}

\usepackage{amsmath,amssymb}
\usepackage{graphicx}
\usepackage{booktabs}
\usepackage{tikz}
\usetikzlibrary{arrows.meta,positioning}

\title{Shared circuits predict whether LLMs generalize \\ across formats in arithmetic reasoning}

\author{
  Andrea Gregor de Varda \quad Sana Pandey \quad Pengrui Han \\
  \textbf{Jacob Andreas}\thanks{\; Equal senior authorship.} \quad
  \textbf{Evelina Fedorenko}\footnotemark[1] \\
  Massachusetts Institute of Technology \\
  \texttt{\{devar\_ag, sanapnde, phan3, jda, evelina9\}@mit.edu}
}

\begin{document}
\maketitle
\begin{abstract}
      In many forms of reasoning, including arithmetic reasoning, generalizing across superficial changes in input format is effortless for humans: anyone who can solve 2+5 can also solve \textit{two plus five}. In contrast, LLMs are more brittle to surface variations of the prompts: for example, they solve numeric arithmetic problems almost perfectly but are substantially less accurate on verbal renditions of the same problems. Here, we ask whether generalization across formats can be predicted from the models' internals. Using attribution patching, we first independently localize the circuit that each model recruits to solve numeric arithmetic problems (2+5) vs. verbal ones, in three languages: English (\textit{two plus five}), Spanish (\textit{dos más cinco}), and Italian (\textit{due più cinque}); then, we test whether overlap with the model's own numeric circuit predicts its generalization to the verbal formats. Indeed, we find support for this idea at three levels: circuit overlap accounts for the relative difficulty of the three verbal formats, for which models generalize best, and for which items are solved correctly, rivaling supervised probes while requiring no labeled data.
\end{abstract}

\section{Introduction}

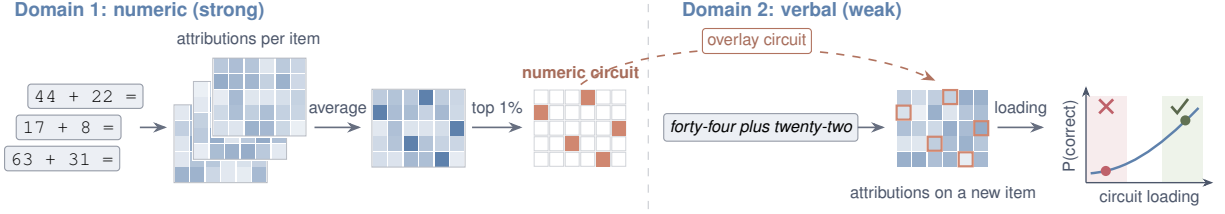
\begin{figure*}
\centering
\resizebox{\linewidth}{!}{\input{figs/fig_pipeline}}
\caption{\textbf{Analysis pipeline.} We first identify the target circuit as the top-1\% of units in the numeric domain (left); then, we use attribution scores on the target circuit in the verbal domain to predict correctness.}
\label{fig:pipeline}
\end{figure*}

\begin{figure*}[!tbp]
\centering
\includegraphics[width=\textwidth]{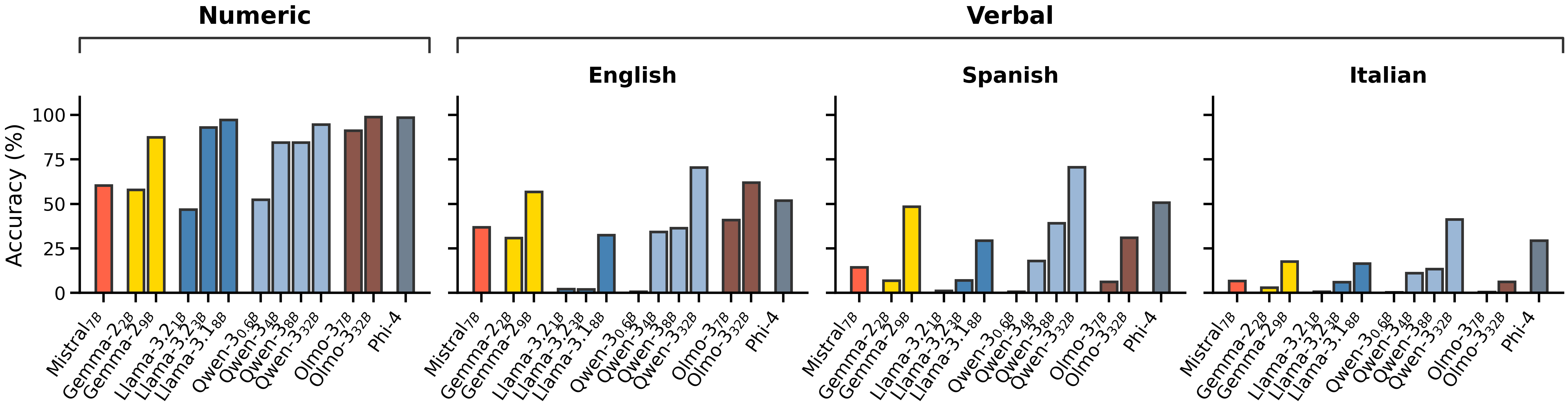}
\caption{\textbf{Per-format accuracy.} Accuracy is high in numeric format and drops in English, Spanish, and Italian.}
\label{fig:accuracy}
\end{figure*}

Whether a model has acquired a capability is, ultimately, a question about generalization: given a set of training examples, can it solve problems that differ from those examples along some relevant dimension? One form of generalization that is particularly easy for humans is generalization across formats. A human who can reliably compute ``$2+5$'' can also answer \textit{``what is the sum of two and five?''} or its translation into another language they know (\textit{``¿Cu\'al es la suma de dos y cinco?''}) without having to learn math again---and this kind of systematicity has long been argued to be a defining property of human cognition \citep{fodor1988,lake2017}.

LLMs, however, are less systematic. They can match or exceed human performance on tasks close to their training distribution, yet fail on minor variations of those same tasks \citep{mccoy2019,mccoy2023}. As LLMs are increasingly deployed in high-stakes settings, it becomes critical to predict how they will perform on new problems---especially problems that deviate from their training distribution---\emph{before} deploying them \citep{yuan2023revisiting}. One approach is to focus on behavior: how well a model does on a target evaluation set. But this approach is limited. First, a given behavior is compatible with infinitely many internal mechanisms \citep{anderson2013adaptive, de2026reply}, so a correct answer is no guarantee that the model arrived at it the ``right'' way. And second, because no benchmark can test all possible variations of a set of problems, a model that solves one variant of a task may fail on another that should be equivalent. To accurately predict generalization then, behavior is not enough. Here, we consider another approach: probing the models' internal mechanisms, which recent work has shown to carry reliable predictive information about behavior \citep{todd2024,panuganti2026,shao2026} and generalization more specifically \citep[][see Appendix~\ref{sec:related} for an extended discussion of related work]{sun2025circuit}. 

Taking arithmetic as a test case---a domain whose internals are well characterized in LLMs \citep{hanna2023,stolfo2023,bertolazzi2025}---we ask whether a model's ability to generalize across surface formats %(e.g., from the numeric format to the verbal format) 
is predictable from the extent to which it relies on the same internal circuit to solve those problems.

For each of 13 LLMs spanning six families and 0.6B--32B parameters, we localize the units recruited by numeric, English, Spanish, and Italian arithmetic, using attribution patching (\citealp{nanda2023,syed2023}; Figure~\ref{fig:pipeline}), and test whether overlap with the model's numeric circuit predicts its verbal format behavior. The prediction holds at three levels: circuit overlap accounts for the relative difficulty of the three verbal formats, for which models generalize best, and for which items are solved correctly, on par with supervised probes despite requiring no labeled data. These findings suggest that there are advantages in considering generalization as not just a property of behavior but of the underlying mechanism. %, and that a model's internals can be used to predict, without any labeled data, where its abilities will and will not transfer.

\section{Methods}

\subsection{Dataset}

Each item in our dataset is an arithmetic problem of the form $a_1\,\square\,a_2\,(\square\,a_3)\,=$ over positive integers, with $\square\in\{+,-\}$. Operands have either 2 or 3 digits and items have either 2 or 3 terms, both balanced 50/50; half of the items require a carry.

Each item is paired with a sign-flipped alternative version, used for attribution patching (§\ref{sec:patching}): every $+$ in the original prompt is replaced by $-$ and vice versa (e.g., original prompt $x = \mbox{\texttt{44 + 22 =}}$ with \say{gold} answer $y = \mbox{\texttt{66}}$; alternative prompt $x' = \mbox{\texttt{44 - 22 =}}$ with gold answer $y' = \mbox{\texttt{22}}$). We use $x, x', y, y'$ in this sense throughout.

Each arithmetic problem is rendered in four formats: \textbf{Numeric} (\texttt{44 + 22 =}), \textbf{English} (\textit{forty-four plus twenty-two equals}), \textbf{Spanish} (\textit{cuarenta y cuatro m\'as veintid\'os es igual a}), and \textbf{Italian} (\textit{quarantaquattro pi\`u ventidue fa}). All analyses use a fixed set of 2,000 items.

\subsection{Models}

We evaluate 13 base language models spanning six families and 0.6\,B--32\,B parameters: Qwen-3 0.6\,B, 4\,B, 8\,B, 32\,B \citep{qwen2025}; Llama-3.2 1\,B and 3\,B, and Llama-3.1 8\,B \citep{llama3}; Gemma-2 2\,B and 9\,B \citep{gemma2024}; Mistral 7\,B \citep{mistral2023}; OLMo-3 7\,B and 32\,B \citep{olmo3}; and Phi-4 14\,B \citep{phi4}. Each model is prompted with a single one-shot in-context exemplar in the target format and decoded greedily. % for $K_i = \max(\text{len}(\text{tok}(y_i)),\,\text{len}(\text{tok}(y'_i)))$ tokens.

\subsection{Attribution patching} \label{sec:patching}

\begin{figure*}[!tbp]
\centering
\includegraphics[width=\textwidth]{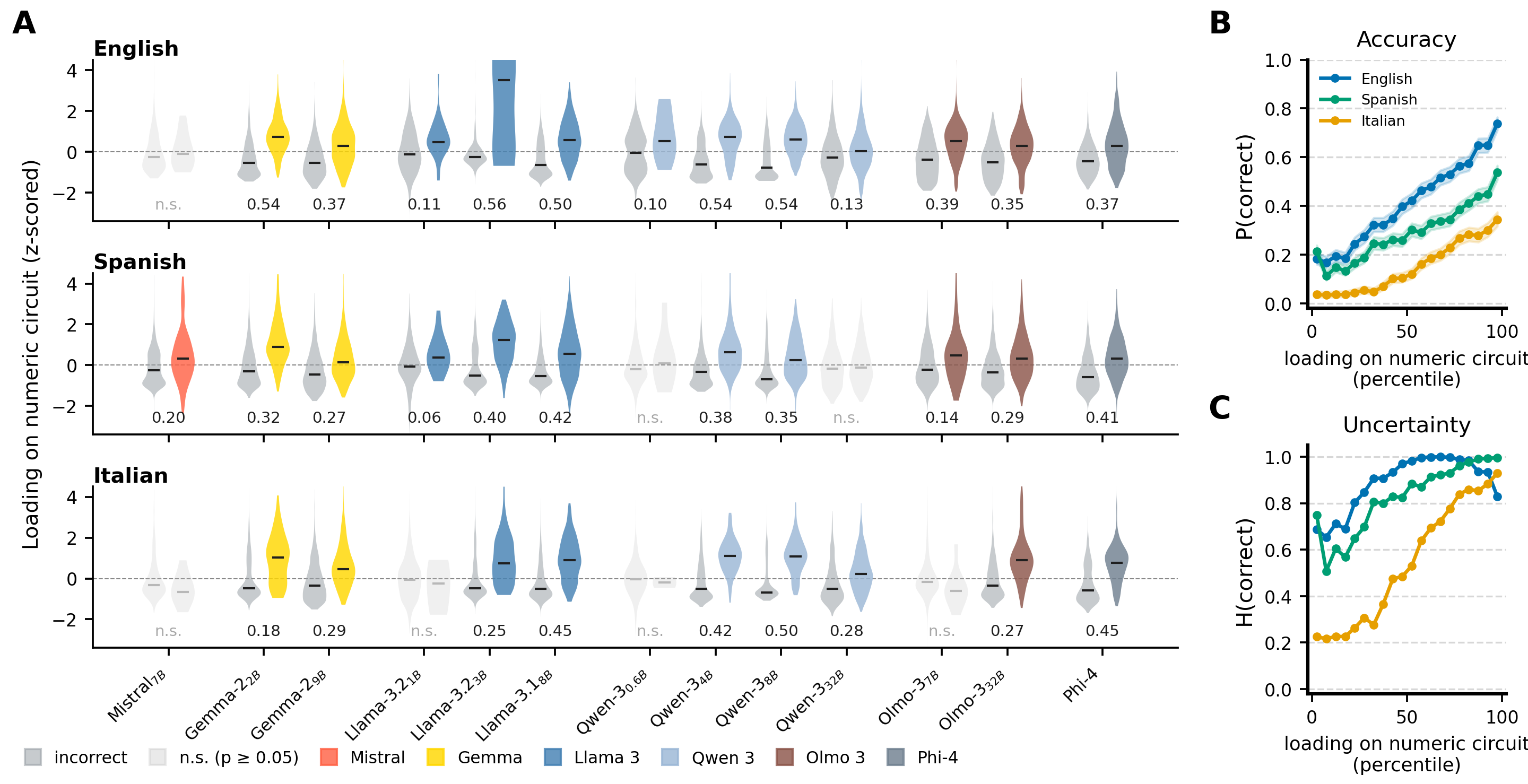}
\caption{\textbf{Circuit engagement predicts item-level accuracy.} \textbf{A} Rows: verbal formats. The paired violins indicate the loading onto the numeric circuit for incorrect (left) and correctly solved items (right) with point-biserial $r$ below; non-significant pairs ($p \ge 0.05$) are in light gray. \textbf{B} Across models, accuracy is near zero at low loading and increases with it. \textbf{C} The uncertainty (entropy) of the outcome also increases with numeric loading.}
\label{fig:violins}
\end{figure*}

Attribution patching (AP; \citealp{nanda2023,syed2023}) derives attribution scores for each unit in a model by linearly approximating the effect of replacing a unit's corrupted activation by its clean value (see Appendix~\ref{sec:ap_details} for details). The standard approach in AP uses pre-specified gold answers, which targets the circuit that \emph{would} produce the right answer rather than the one the model actually uses---a problem in formats where the model is mostly wrong. 

We instead derive attribution scores from the model's own answers. We take an arithmetic prompt and its sign-flipped version (e.g. \texttt{44 + 22 =} and \texttt{44 - 22 =}), record the model's greedy answer to each, and ask how strongly the model favors the plus-answer over the minus-answer; we then localize the units responsible for that preference. Because it only uses the model's own outputs and never a gold answer, the metric is well-defined even when the model is usually wrong. We rescale this preference to 0 on the sign-flipped prompt and 1 on the original so attributions can be averaged and compared across formats.
We drop items in which the clean and counterfactual prompts tokenize to different lengths, or in which the model's outputs in response to $x$ and $x'$ are identical.

For each model $m$ and format $f$ we average the item-level unit attribution scores to obtain a format-level score  $A^{m,f}\in\mathbb{R}^{L\times d_{\mathrm{MLP}}}$ and define the \textbf{numeric circuit} $S^m$ as the top 1\% of units of $A^{m,\text{numeric}}$ by attribution scores. For any verbal format $f$ we then report two quantities relative to $S^m$: a) the circuit's \textbf{overlap} with the numeric circuit (top-1\% Jaccard between $A^{m,f}$ and $A^{m,\text{numeric}}$), and b) an item's \textbf{loading} on the numeric circuit (the sum of attribution scores on units in $S^m$, $\mathrm{loading}_i = \sum_{(L,u)\,\in\,S^m} \mathrm{attrib}_{i,L,u}$, where $i$ is an item; see Figure~\ref{fig:pipeline}).

\section{Results}

\subsection{Accuracy varies dramatically with format}

All 13 models solve numeric arithmetic well above chance (median 87.4\%) %range 46.9\,--\,98.8\%)
but drop substantially in English (36.4\%), further in Spanish (17.9\%), and lowest in Italian (6.8\%; Figure~\ref{fig:accuracy}). The ordering Numeric $>$ English $\geq$ Spanish $\geq$ Italian holds in every model. This pattern of performance across formats is the variability we attempt to capture at the level of circuits.

\subsection{Circuit overlap predicts format- and model-level accuracy}

% \begin{figure}[!tbp]
% \centering
% \includegraphics[width=\columnwidth]{figs/fig2_format_overlap.png}
% \caption{Format-level overlap with the numeric circuit (left) and item-level accuracy (right) share the same English $>$ Spanish $>$ Italian profile. Bars indicate cross-model means; dots indicate individual models.}
% \label{fig:overlap}
% \end{figure}

For each model we computed the top-1\% Jaccard between each verbal format circuit and the same model's numeric circuit. On average, more units are shared between the numeric circuit and the English circuit ($0.145$), followed by Spanish ($0.078$) and Italian ($0.069$). This pattern held for each individual model in our sample. %(Figure~\ref{fig:overlap}). 
The verbal formats whose circuit aligns most with the numeric circuit are also the formats LLMs are most accurate at.

%\subsection{Overlap predicts model-level accuracy}

Across models, overlap with each model's own numeric circuit predicts that model's accuracy in English ($r=0.75$, $p=0.003$, $N = 13$) and Spanish ($r=0.58$, $p=0.04$)---although the trend is not significant in Italian ($r=0.47$, $p=0.10$).%, Figure~\ref{fig:scatter}).

% \begin{figure}[h]
% \centering
% \includegraphics[width=\columnwidth]{figs/fig3_model_scatter_compact.png}
% \caption{Cross-model overlap with the numeric circuit (\textit{x}-axis) predicts accuracy in the matching verbal format (\textit{y}-axis). }
% \label{fig:scatter}
% \end{figure}

\subsection{Circuit loading predicts item accuracy}
\label{sec:main-item-loading}

Even models with low verbal accuracy get some items right; we asked whether we could predict item-level accuracy based on whether the models used the numeric circuit to solve that problem. For each combination of model and format, we computed the loading onto the numeric circuit for each item, and tested its association with accuracy using point-biserial correlation. In every combination with at least 30 correct items, correct items loaded onto the numeric circuit more than incorrect items (Figure~\ref{fig:violins}A). The point-biserial $r$ is significant and positive ($p<0.05$) for 12/13 models in English (range $0.11$ to $0.56$, median $0.38$), 11/13 in Spanish ($0.06$ to $0.42$, median $0.32$), and 9/13 in Italian ($0.19$ to $0.50$, median $0.29$). %Non-significant combinations belong to the smallest models (Qwen-3$_{0.6\mathrm{B}}$, Llama-3.2$_{1\mathrm{B}}$) in the lowest-accuracy formats, where the number of correct items is very small. Loading onto the numeric circuit is a particularly strong predictor of accuracy for Llama-3.1-8B, with $r_{\mathrm{pb}}=0.50$ (English), $0.42$ (Spanish), $0.45$ (Italian), all $p<0.0001$. 
Across models, at low loading on the numeric circuit, accuracy is near zero; as loading increases, accuracy increases as well (Figure~\ref{fig:violins}B), but so does the entropy of correctness (Figure~\ref{fig:violins}C). In other words, if the numeric circuit is not engaged, then the response in the verbal formats will almost definitely be incorrect, but if the numeric circuit is engaged, the outcome is more likely to be correct but is still uncertain and models may fail for other reasons.

These results are not an artifact of the linear approximation in attribution
patching: patching activations in the numeric circuit units confirms
that the circuit's causal involvement predicts whether the item is solved
(Appendix~\ref{sec:causal}). The fact that we can use the numeric circuit to predict correctness in other formats is tied to the high accuracy of LLMs in numeric arithmetic: if circuits are localized on numeric problems that the models are not able to solve, those circuits are not predictive of cross-format generalization (Appendix~\ref{sec:control}). In fact, we show that any format can serve as the \textit{source} format to predict performance in all other \textit{target} formats (e.g., using circuits identified on Italian problems to predict accuracy on numeric problems), as long as we balance the accuracy of our problem set (Appendix~\ref{sec:allpairs}). Yet, using any of the verbal formats as the source is impractical, given that accuracy in those formats is generally low. %, and to obtain a sample of correctly-solved items requires searching over a large space of problems. 

\subsection{Circuit loading predicts accuracy beyond probes and confidence}
\label{sec:ensemble}

\begin{figure}[!tbp]
\centering
\includegraphics[width=\linewidth]{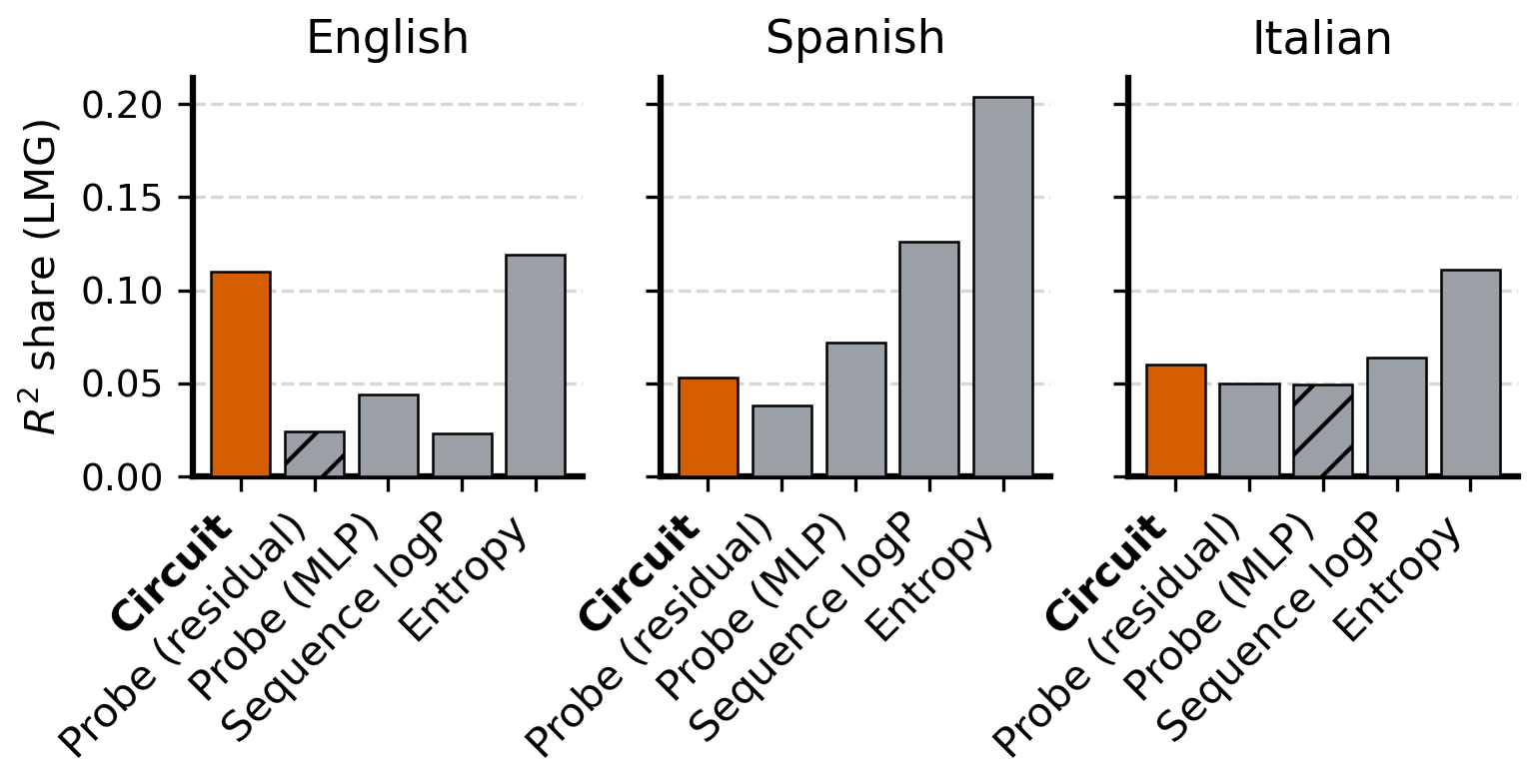}
\caption{\textbf{Circuit loading predicts item-level accuracy beyond trained probes and confidence measures.} Share of the variance in item-level correctness attributed to each predictor by an $R^2$ decomposition. Hatched bars indicate predictors that are not significant ($p\geq0.05$).}
\label{fig:variance_partition}
\end{figure}

To test whether our circuit loading metric is redundant with simpler correlates of success, we compared it against four controls on Llama-3.1-8B: two supervised probes trained on held-out numeric problems to predict correctness from internal representations (one on the residual stream, one on the MLP activations), and two confidence measures derived from the output distribution (the mean log-probability of the model's own answer, and next-token entropy at the decision point; see Appendix~\ref{sec:ensemble_details} for details). We entered the five predictors in a linear probability model of item-level correctness and decomposed its $R^2$. Circuit loading remained significant in all three formats with all controls in the model ($\beta=0.127$, $p<0.0001$ in English; $\beta=0.030$, $p=0.002$ in Spanish; $\beta=0.031$, $p=0.006$ in Italian; Figure~\ref{fig:variance_partition}). In English, it accounted for as much variance as answer entropy, the strongest control (11.0\% vs.\ 11.9\%); in Spanish and Italian, entropy
was stronger (5.3\% vs.\ 20.4\%, and 6.0\% vs.\ 11.1\%). In all formats, circuit loading matched or exceeded both trained probes. %, despite requiring no labeled data.

\section{Discussion}

We found that circuits predict generalization in arithmetic reasoning performance at three levels: they account for the relative difficulty of the three verbal formats, for which models generalize best, and for which items are solved correctly. In short, the generalization of behavior is grounded in the generalization of the circuit: when the circuit does not carry over to a new format, arithmetic abilities do not carry over either. Furthermore, the relationship is asymmetric: if a model does not engage the numeric circuit it almost always fails to generalize to verbal formats, whereas if a model engages the circuit it will not be guaranteed to generalize. We can thus predict failure more confidently than success. These results extend prior work linking circuits to generalization---most closely, \citet{sun2025circuit}, who predict aggregate performance from circuit stability---to changes in surface format and to predictions about individual problems.

Our approach is general and can therefore be straightforwardly extended to other aspects of mathematical reasoning or to other domains of reasoning. The approach is especially useful when there are two or more problem formats with asymmetric performance: the circuit is localized using the format with high performance, and its engagement is measured in the format with lower performance. Such asymmetries in LLMs' abilities are common: models generate better code in high-resource programming languages such as Python than in low-resource ones like OCaml or Racket \citep{cassano2023multipl} and answer the same questions more accurately in English than in low-resource languages \citep{ahuja2023mega}. In each case, the strong format can be used to identify the reference circuit for a target ability to predict generalization. % in the weaker formats.

Critically, our method requires no supervision or ground truth, 
unlike other techniques like probing. The only prior knowledge needed is which format is the strong one, and this is usually known in advance. And yet, our circuit metric predicts generalization on par or better than probes trained on thousands of examples.

\section*{Limitations}

This paper presents an empirical finding, but at the moment, we do not have a theory that can explain the success of our approach. We can show that circuit overlap and behavioral generalization go together across formats, models, and items, but we cannot yet say why, or under which conditions the relationship should hold or break. We hope this work can serve as an empirical basis for developing a theory of generalization grounded in the models' internal computations.

A related limitation is that our circuits are sets of units engaged by the problems, and we do not directly link them to the algorithms they implement. In other words, we can measure whether two formats engage the same units, but we cannot infer the computations that those units support. Connecting the implementation level (which components a model uses) to the algorithmic level (what procedure those components carry out; \citealp{marr1982}) would be necessary to move from statistical prediction, which is what we have now, to formal guarantees of correctness, which is the ultimate requirement for deploying LLMs in high-stakes settings. We are admittedly far from that point, but we hope our findings are a step toward it.

Lastly, our method explains significant variance above and beyond existing techniques, but it is not the single strongest predictor of correctness: entropy at the decision point, which also requires no ground-truth data, explains as much or more variance than the loading on the numeric circuit, clearly more so in the lower-resource formats. Still, the two signals are not interchangeable but complementary, as circuit loading contributes significant unique variance in all three formats. Additionally, entropy only indicates that a model is uncertain, whereas circuit loading links that uncertainty to the model's internal structure.

\section*{Ethical Considerations}
This work is fundamental interpretability research on arithmetic reasoning using synthetic data and publicly available models; we do not foresee ethical, societal, or environmental risks beyond those of the underlying models. We used Claude Opus 4.5--4.8 for coding and writing assistance; all scientific content, design decisions, and claims are the authors' own.

\section*{Acknowledgements}
We are grateful to members of the DARPA AIQ team for comments. AGdV was supported by the K. Lisa Yang ICoN Center Postdoctoral Fellowship. SP was supported by the National Science Foundation Graduate Research Fellowship Program under Grant No.\ 2141064. JA was supported by research funds from the MIT Siegel Family Quest for Intelligence. EF was supported by research funds from the McGovern Institute for Brain Research, the Simons Center for the Social Brain, and the MIT Siegel Family Quest for Intelligence. JA and EF and this research were partially supported by the Defense Advanced Research Projects Agency (DARPA) AIQ program through the DARPA CMO contract number HR00112520025. Any opinions, findings, and conclusions or recommendations expressed in this material are those of the author(s) and do not necessarily reflect the views of the National Science Foundation or DARPA.

%%%%%%%%%%%%%%%%%%%%%%%%%%%%%%%%%%%%%%%%%%%%%%%%%%%%%%%%%%%%%%%%%%%%%%%%%%%%
%%%%%%%%%%%%%%%%%%%%%%%%%%%%%%%%%%%%%%%%%%%%%%%%%%%%%%%%%%%%%%%%%%%%%%%%%%%%
%%%%%%%%%%%%%%%%%%%%%%%%%%%%%%%%%%%%%%%%%%%%%%%%%%%%%%%%%%%%%%%%%%%%%%%%%%%%

\bibliography{references}

\appendix

\section{Supplementary related work}
\label{sec:related}
\paragraph{Circuits are reused across tasks and languages.} A single circuit often supports different behaviors \citep{merullo2024,nainani2024}, and circuits are reused across languages in multilingual models \citep{ferrando2024,wendler2024,lindsey2025}. Most relevant to our approach is the finding from \citet{lan2024} that a shared circuit represents digits and Spanish number words.

\paragraph{Arithmetic circuits.} Numeric arithmetic has well-characterized internals: a greater-than circuit in GPT-2 \citep{hanna2023}, a three-stage addition process in base LMs \citep{stolfo2023}, and \say{consistency} heads that check numerical alignment \citep{bertolazzi2025}. \citet{wu2025semantic} further show that LLMs map numeric and verbal arithmetic expressions into a shared intermediate space.

\paragraph{Circuits as predictors.} Few studies have asked whether signals derived from circuits predict model performance \citep{todd2024,panuganti2026,shao2026}. Most related, \citet{sun2025circuit} uses circuit stability across arithmetic subtasks to predict generalization, but varies problem structure within a single numeric format and predicts accuracy at the subtask level, not at the item level. To our knowledge, no previous study used an independently localized reference circuit to predict generalization across formats, and none has made predictions at the level of individual problems.

\section{Attribution patching}
\label{sec:ap_details}
For each item we have the original prompt $x$, its sign-flipped version $x'$, and the model's greedily decoded answers to each, $\hat{y}$ and $\hat{y}'$. We define a preference score that measures how strongly the model favors $\hat{y}$ over $\hat{y}'$ when reading an input $z$:
\begin{equation}
m(z) = \log P(\hat{y} \mid z) - \log P(\hat{y}' \mid z)
\end{equation}
where each log-probability is the teacher-forced sum over answer tokens. By construction, $m$ is high when $z = x$ and low when $z = x'$. We rescale it so that it equals 1 on the original prompt and 0 on the sign-flipped one:
\begin{equation}
M(z) = \frac{m(z) - m(x')}{m(x) - m(x')}
\end{equation}
No gold answer is used to produce $m$ or $M$: both are constructed from the model's own responses.
Activation patching would replace, while the model processes $x'$, the activation $a_i$ of a single unit $i$ with the value it takes on $x$, and record the change in $M$; units whose replacement has a substantial impact on $M$ are the most important units for the task. Doing this exhaustively requires one forward pass per unit, which is intractable for large models. Attribution patching linearly approximates the effect of every such replacement at once, assigning each unit $i$ an attribution score:
\begin{equation}
\mathrm{attrib}_i = \left(a_i(x) - a_i(x')\right)\, \nabla_i M
\end{equation}
where $a_i(x)$ and $a_i(x')$ are the activations of unit $i$ when the model reads the original prompt $x$ and the alternative $x'$, and $\nabla_i M$ is the gradient of the metric $M$ with respect to unit $i$, evaluated on the alternative prompt $x'$.
The units we score are the MLP units of each layer read at the last position in the prompt (i.e., the position whose logits produce the first answer token). For each item, the procedure require two forward passes ($x$ and $x'$) and one backward pass (on $x'$), and returns an attribution score for every unit in the model. We base our implementation on that of \citet{han2026modular}.

\section{Causal validation of the item-level prediction}
\label{sec:causal}

Attribution patching linearly approximates the effect of an intervention that swaps activations between two versions of a prompt. Here we ran the intervention itself on all 13 models with the main item set, to test whether the item-level prediction of correctness remains when the approximation is removed.

In the intervention, the model reads the sign-flipped alternative prompt $x'$, but at the last prompt token we overwrite the activations of the units in the numeric circuit $S^m$ with the values they take when the model reads the original prompt $x$. If these units carry the relevant computation for an item, the patch should shift the model's preference away from the sign-flipped answer and back toward the original one. We quantify this shift as the fraction of the preference difference that the patch restores: 0 means the patch had no effect, 1 means the preference was fully restored to its value on the original prompt. As a control, we repeat the procedure on a random set of units, matched in number to $S^m$ within each layer.

\begin{figure}[!tbp]
\centering
\includegraphics[width=\columnwidth]{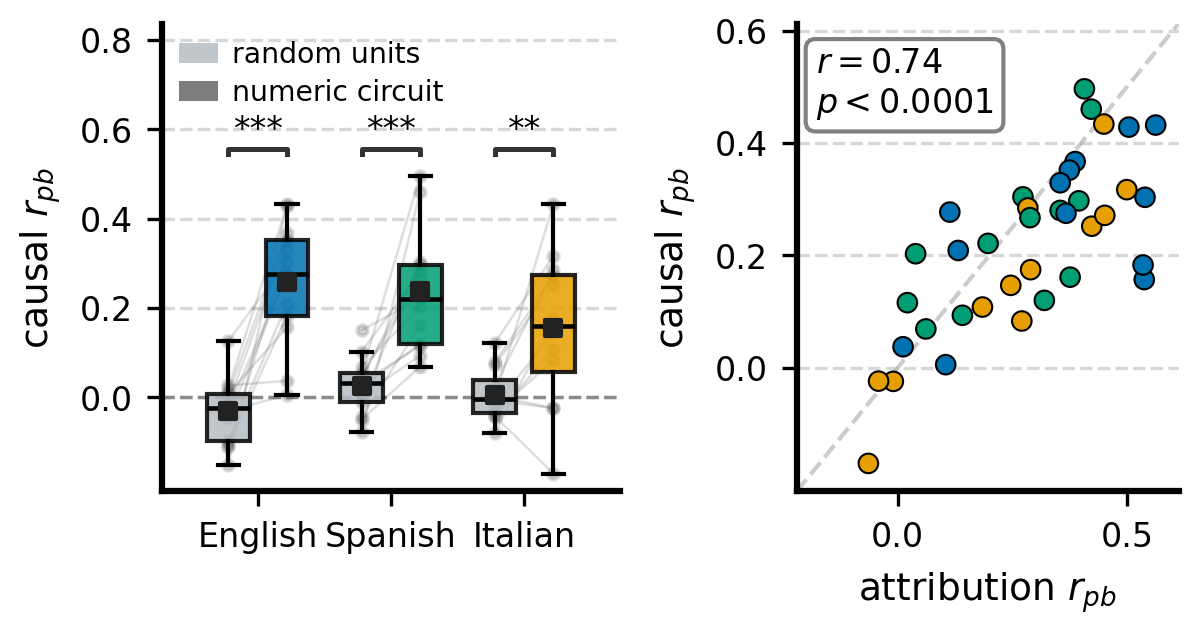}
\caption{\textbf{The numeric circuit causally drives item-level accuracy.}
\textbf{A} Point-biserial correlation between the fraction of preference restored by activation patching and item-level accuracy for random units (gray) and the numeric circuit (colored). Each faint line is one model, squares are means, and brackets are paired Wilcoxon tests ($p<0.01$**, $p<0.001$***). \textbf{B} Across models and formats, the causal correlation (\textit{y}) correlates with the attribution-based correlation of the main analysis (\textit{x}).}
\label{fig:causal}
\end{figure}

Across models, the items on which patching restores more of the preference (that is, the items where the numeric circuit has the strongest causal effect) are the items the model answers correctly. The point-biserial correlation between the restored fraction and accuracy is positive in 35 of the 38 model and format combinations, and significant in 32. The correlation is much larger for the numeric circuit than for random units in every format (Figure~\ref{fig:causal}A; paired Wilcoxon $p<0.001$ in English and Spanish, $p<0.01$ in Italian). Moreover, the predictivity of the circuit-based estimates obtained with the causal and the attribution-based approach are correlated across models and formats ($r=0.74$, Figure~\ref{fig:causal}B). The item-level results of the main text (§\ref{sec:main-item-loading}) therefore reflect a causal dependency on the numeric circuit.

\section{Probe and confidence-measure details}
\label{sec:ensemble_details}

The probes of §\ref{sec:ensemble} are logistic regressions that read the hidden state at the last prompt token and predict whether the model will answer correctly. One probe reads the residual stream; the other reads the same MLP activations that attribution patching scores. Both were trained on 2,000 numeric problems held out from the main item set (1,600 correct,
400 incorrect; see Appendix \ref{sec:control} for details on the data); class weights were used to mitigate class imbalance. The layer and the regularization strength were selected with stratified 5-fold cross-validation. Both probes selected layer 25 and reached a cross-validated AUC of 0.93 (residual stream) and 0.92 (MLP activations) on numeric problems. Like the circuit, the probes are estimated on numeric arithmetic and transferred to the verbal formats; unlike the circuit, they are explicitly fitted to predict correctness.

The two confidence measures are derived from the model's output distribution. The mean log-probability is the teacher-forced average over the tokens of the model's own answer; the entropy is computed over the next-token distribution at the last token of the prompt (i.e., at the decision point before producing the answer).

We entered the five predictors in a linear probability model of item-level correctness and decomposed its $R^2$ with the LMG method. LMG relative importance (named after Lindemann, Merenda and Gold, the creators of the metric; \citealp{lindeman1980,gromping2006}) averages each predictor's contribution to the $R^2$ over all orders in which predictors can enter the model.

\section{Disentangling correctness and format}
\label{sec:control}

The main result is that loading on the numeric circuit predicts whether an item presented in a verbal format is solved correctly. Here, we test whether this depends on the fact that the numeric circuit is identified from a sample where the model is correct $\sim$97\% of the time, whereas verbal aggregates come from samples where the model is correct on only 17\,--\,38\% of items. In other words, we test if the numeric circuit predicts accuracy in other formats \textit{because} it obtains high accuracy in general. We test this on Llama-3.1-8B.

We generated a new dataset of 20,000 items, computed item-level accuracy across formats, and then sampled per format independently 1,600 correct and 400 incorrect items, giving 2,000 items per format at an 80/20 correct/incorrect split. We built three variants of the numeric circuit on the matched numeric sample: $S^{\text{all}}$ (all 2,000 items), $S^{\text{correct}}$ (1,600 correct), $S^{\text{incorrect}}$ (400 incorrect). For each verbal format we then computed the items' loading on each variant of the numeric circuit and its point-biserial correlation with accuracy.

Loadings on $S^{\text{all}}$ and $S^{\text{correct}}$ predict verbal-format accuracy at essentially the same magnitude (Table~\ref{tab:control_rpb}); loadings on $S^{\text{incorrect}}$ do not predict accuracy across formats. 

\begin{table}[h]
\centering\small
\begin{tabular}{lccc}
\toprule
verbal & $S^{\text{all}}$ & $S^{\text{correct}}$ & $S^{\text{incorrect}}$ \\
\midrule
English & $0.38$ & $0.38$ & $-0.05$ \\
Spanish & $0.30$ & $0.30$ & $0.11$ \\
Italian & $0.38$ & $0.38$ & $0.10$ \\
\bottomrule
\end{tabular}
\caption{Item-level point-biserial correlation between loading on the numeric circuit and accuracy, under three definitions of the numeric circuit. $n=2{,}000$ per format. Results from Llama-3.1-8B.}
\label{tab:control_rpb}
\end{table}

\section{Cross-format prediction for all format pairs}
\label{sec:allpairs}

\begin{figure}[!tbp]
\centering
\includegraphics[width=\columnwidth]{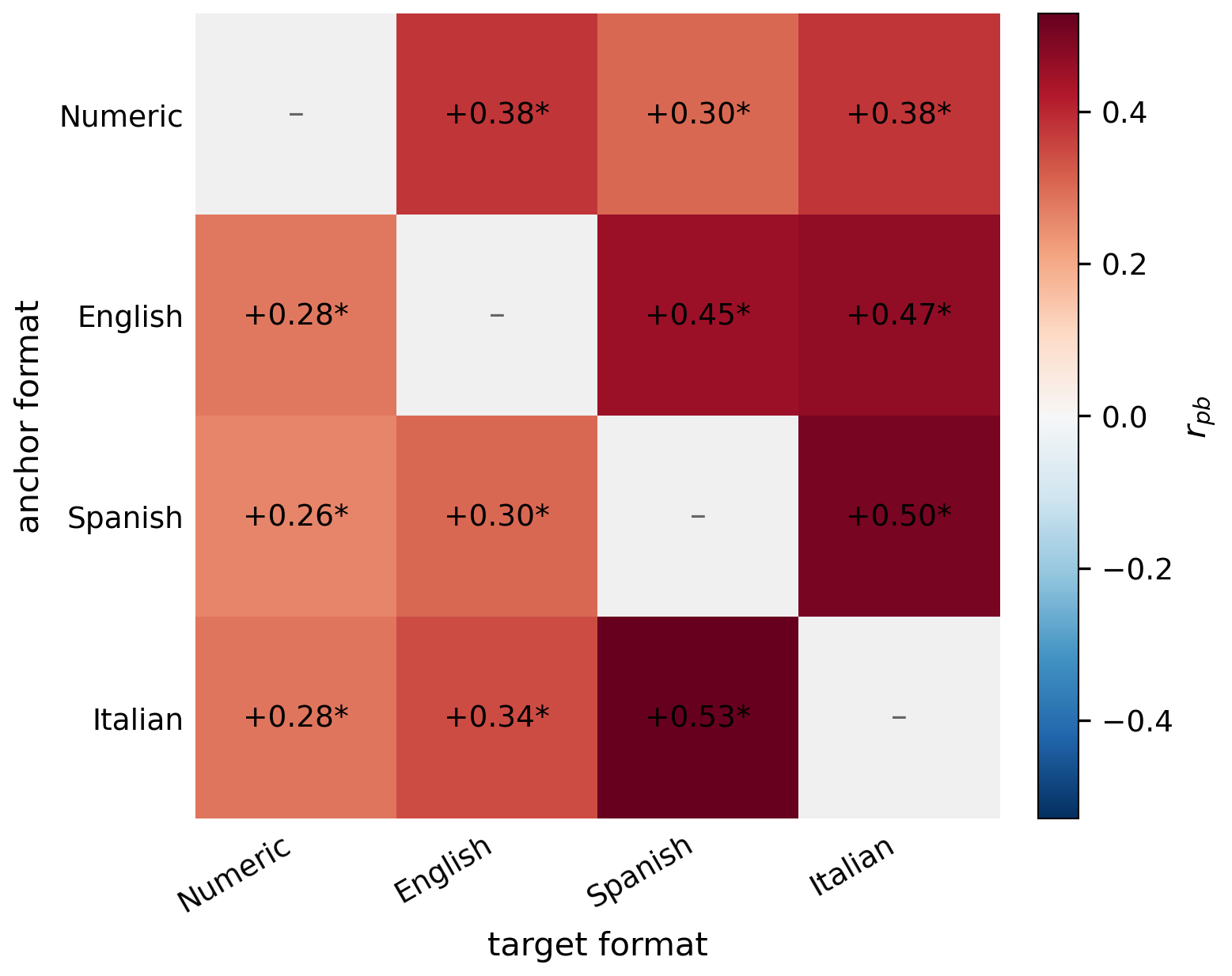}
\caption{\textbf{Every format's circuit predicts item-level accuracy in every other format.} Point-biserial correlation between loading on the circuit identified in the anchor format (rows) and item-level correctness in the target format (columns), computed on the balanced samples ($n=2{,}000$ per format, 80\% correct). Asterisks mark significance at $p<0.05$ (Holm-corrected). Results from Llama-3.1-8B.}
\label{fig:allpairs_balanced}
\end{figure}

The main analyses use the numeric format as the \textit{anchor} format where the reference circuit is identified. The numeric format is the only format where models achieve high accuracy (median accuracy 87.4\%), so the numeric circuit is identified from mostly correct behavior. Here we ask whether the numeric circuit is privileged in predicting accuracy across formats or if instead any format can be used as long as we use a problem sample with high accuracy. To do so, we repeat the item-level analysis for all 16 anchor--target format pairs on Llama-3.1-8B.

Accuracy differs strongly across formats, so if we identified circuits on the raw samples used in the main analysis we would confound format with accuracy. We therefore reuse the balanced design of Appendix~\ref{sec:control}: for each format, 1,600 correct and 400 incorrect items ($n=2{,}000$, accuracy fixed at 80\%). For each anchor format we define the circuit as the top 1\% of units of that format's balanced aggregate. For each target format we compute each item's loading on the anchor circuit and correlate it with correctness.

Every anchor predicts every target ($r_{\mathrm{pb}}=0.26$ to $0.53$, all Holm-corrected $p<0.05$; Figure~\ref{fig:allpairs_balanced}). Verbal anchors predict verbal targets most strongly ($0.30$ to $0.53$); the numeric anchor predicts verbal targets at $0.30$ to $0.38$, and verbal anchors predict numeric accuracy at $0.26$ to $0.28$. Thus, numeric circuits predict verbal accuracy better than the other way around, but cross-format prediction is not specific to the numeric circuit: any format's circuit predicts accuracy in any other format, as expected if all formats tap onto a shared arithmetic circuit to different degrees.

\end{document}

%% file: figs/fig_pipeline.tex
% Pipeline figure body (wide single-band layout for figure*).
% In main.tex preamble add:
%   \usepackage{tikz}\usetikzlibrary{arrows.meta,positioning}
% then use inside a figure* environment:
%   \resizebox{\textwidth}{!}{\input{fig_pipeline_body}}
\definecolor{PBlue}{HTML}{5E81AC}%
\definecolor{POrange}{HTML}{D08770}%
\definecolor{PRed}{HTML}{BF616A}%
\definecolor{PGreen}{HTML}{A3BE8C}%
\definecolor{PGray}{HTML}{4C566A}%
\definecolor{PBg}{HTML}{ECEFF4}%
\begin{tikzpicture}[x=1mm,y=1mm, font=\sffamily,
    card/.style={draw=PGray!60, rounded corners=1.2pt, fill=PBg,
                 inner sep=2.2pt, font=\sffamily\scriptsize\ttfamily, text=black},
    lab/.style={font=\sffamily\tiny, text=PGray},
    biglab/.style={font=\sffamily\scriptsize\bfseries, text=PBlue},
    arr/.style={-{Stealth[length=1.8mm]}, semithick, draw=PGray!85}]

  \def\cs{2.0} % cell size; grids are 6 x 5 cells (12mm x 10mm)

  % ================= Domain 1 =================
  \node[biglab, anchor=west] at (0,24.5) {Domain 1: numeric (strong)};

  % cascaded prompt cards
  \node[card, anchor=west, minimum width=13.5mm] at (0,5.5) {63 + 31 =};
  \node[card, anchor=west, minimum width=13.5mm] at (1.5,9.5) {17 + 8 =};
  \node[card, anchor=west, minimum width=13.5mm] at (3,13.5) {44 + 22 =};

  % cascaded attribution grids
  \foreach \gx/\gy/\seed in {22/2.5/7, 24.5/5.5/13, 27/8.5/42}{
    \pgfmathsetseed{\seed}
    \fill[white] (\gx-0.3,\gy-0.3) rectangle (\gx+6*\cs,\gy+5*\cs);
    \foreach \i in {0,...,5}{\foreach \j in {0,...,4}{
      \pgfmathsetmacro{\op}{12+55*rnd}
      \fill[PBlue!\op] ({\gx+\i*\cs},{\gy+\j*\cs}) rectangle ({\gx+\i*\cs+1.8},{\gy+\j*\cs+1.8});
    }}
    \draw[PGray!40, line width=0.3pt] (\gx-0.1,\gy-0.1) rectangle (\gx+6*\cs-0.1,\gy+5*\cs-0.1);
  }
  \draw[arr] (17.5,9.5) -- (21,9.5);
  \node[lab] at (31.5,20.8) {attributions per item};

  % merged grid
  \draw[arr] (40,9.5) -- (46,9.5) node[midway, above, lab] {average};
  \pgfmathsetseed{42}
  \foreach \i in {0,...,5}{\foreach \j in {0,...,4}{
    \pgfmathsetmacro{\op}{15+45*rnd}
    \fill[PBlue!\op] ({47.5+\i*\cs},{4.5+\j*\cs}) rectangle ({47.5+\i*\cs+1.8},{4.5+\j*\cs+1.8});
  }}
  \foreach \i/\j in {0/3,2/1,3/4,4/0,5/2}{
    \fill[PBlue] ({47.5+\i*\cs},{4.5+\j*\cs}) rectangle ({47.5+\i*\cs+1.8},{4.5+\j*\cs+1.8});
  }
  \draw[PGray!40, line width=0.3pt] (47.4,4.4) rectangle (59.4,14.4);

  % threshold -> circuit mask
  \draw[arr] (60.3,9.5) -- (66.8,9.5) node[midway, above, lab] {top 1\%};
  \foreach \i in {0,...,5}{\foreach \j in {0,...,4}{
    \draw[PGray!25, line width=0.3pt] ({68.3+\i*\cs},{4.5+\j*\cs}) rectangle ({68.3+\i*\cs+1.8},{4.5+\j*\cs+1.8});
  }}
  \foreach \i/\j in {0/3,2/1,3/4,4/0,5/2}{
    \fill[POrange] ({68.3+\i*\cs},{4.5+\j*\cs}) rectangle ({68.3+\i*\cs+1.8},{4.5+\j*\cs+1.8});
  }
  \node[font=\sffamily\tiny\bfseries, text=POrange!80!black] at (74.3,16.6) {numeric circuit};

  % light separator between domains
  \draw[PGray!30, dashed, line width=0.4pt] (83,-1) -- (83,26);

  % ================= Domain 2 =================
  \node[biglab, anchor=west] at (86,24.5) {Domain 2: verbal (weak)};

  \node[card, font=\sffamily\tiny\itshape] at (97.5,9.5) {forty-four plus twenty-two};

  \draw[arr] (110,9.5) -- (113.5,9.5);

  % attribution grid for verbal item, with circuit overlay
  \pgfmathsetseed{99}
  \foreach \i in {0,...,5}{\foreach \j in {0,...,4}{
    \pgfmathsetmacro{\op}{12+55*rnd}
    \fill[PBlue!\op] ({115+\i*\cs},{4.5+\j*\cs}) rectangle ({115+\i*\cs+1.8},{4.5+\j*\cs+1.8});
  }}
  \foreach \i/\j in {0/3,2/1,3/4,4/0,5/2}{
    \draw[POrange, line width=0.8pt] ({115+\i*\cs},{4.5+\j*\cs}) rectangle ({115+\i*\cs+1.8},{4.5+\j*\cs+1.8});
  }
  \node[lab] at (121,1.2) {attributions on a new item};

  % overlay arrow from circuit mask over to verbal grid, boxed label
  \draw[arr, dashed, draw=POrange!85!black] (74.3,15) .. controls (82,21) and (108,21) .. (120,15.5);
  \node[draw=POrange!85!black, fill=white, rounded corners=1.2pt, inner sep=1.8pt,
        font=\sffamily\tiny, text=POrange!80!black] at (97,20.6) {overlay circuit};

  % ---- loading -> performance mini plot ----
  \draw[arr] (128,9.5) -- (134,9.5) node[midway, above, lab] {loading};

  \begin{scope}[shift={(139,2)}]
    % shaded regions
    \fill[PRed!12]   (0.4,0.4) rectangle (5.6,11.5);
    \fill[PGreen!18] (10.2,0.4) rectangle (15.4,11.5);
    % axis lines
    \draw[PGray, line width=0.5pt, -{Stealth[length=1.4mm]}] (0.4,0.4) -- (16.8,0.4);
    \draw[PGray, line width=0.5pt, -{Stealth[length=1.4mm]}] (0.4,0.4) -- (0.4,12.8);
    % rising curve
    \draw[PBlue, line width=0.9pt]
      (1.0,1.6) .. controls (5.5,2.0) and (9.4,4.4) .. (14.9,10.0);
    % example items on the curve
    \fill[PRed]   (2.95,1.9) circle (0.65);
    \fill[PGreen!60!black] (13.2,8.4) circle (0.65);
    % zone marks: cross and check at the same height
    \draw[PRed, line width=1.0pt] (2.1,9.3) -- (3.9,11.1);
    \draw[PRed, line width=1.0pt] (2.1,11.1) -- (3.9,9.3);
    \draw[PGreen!55!black, line width=1.0pt] (11.4,10.2) -- (12.15,9.35) -- (13.55,11.15);
    % axis labels
    \node[lab] at (8.6,-1.6) {circuit loading};
    \node[lab, rotate=90] at (-1.6,6.4) {P(correct)};
  \end{scope}
\end{tikzpicture}%